# Asymmetric Distributed Constraint Optimization Problems

**Tal Grinshpoun**                                              GRINSHPO@CS.BGU.AC.IL
**Alon Grubshtein**                                             ALONGRUB@CS.BGU.AC.IL
*Department of Computer Science*
*Ben-Gurion University of the Negev*
*Beer-Sheva, Israel*

**Roie Zivan**                                                  ZIVANR@BGU.AC.IL
*Department of Industrial Engineering and Management*
*Ben-Gurion University of the Negev*
*Beer-Sheva, Israel*

**Arnon Netzer**                                                NETZERAR@CS.BGU.AC.IL
**Amnon Meisels**                                               AM@CS.BGU.AC.IL
*Department of Computer Science*
*Ben-Gurion University of the Negev*
*Beer-Sheva, Israel*

## Abstract

Distributed Constraint Optimization (DCOP) is a powerful framework for representing and solving distributed combinatorial problems, where the variables of the problem are owned by different agents. Many multi-agent problems include constraints that produce *different gains* (or costs) for the participating agents. *Asymmetric* gains of constrained agents cannot be naturally represented by the standard DCOP model.

The present paper proposes a general framework for Asymmetric DCOPs (ADCOPs). In ADCOPs different agents may have different valuations for constraints that they are involved in. The new framework bridges the gap between multi-agent problems which tend to have asymmetric structure and the standard symmetric DCOP model. The benefits of the proposed model over previous attempts to generalize the DCOP model are discussed and evaluated.

Innovative algorithms that apply to the special properties of the proposed ADCOP model are presented in detail. These include complete algorithms that have a substantial advantage in terms of runtime and network load over existing algorithms (for standard DCOPs) which use alternative representations. Moreover, standard incomplete algorithms (i.e., local search algorithms) are inapplicable to the existing DCOP representations of asymmetric constraints and when they are applied to the new ADCOP framework they often fail to converge to a local optimum and yield poor results. The local search algorithms proposed in the present paper converge to high quality solutions. The experimental evidence that is presented reveals that the proposed local search algorithms for ADCOPs achieve high quality solutions while preserving a high level of privacy.

## 1. Introduction

*"The universe is asymmetric and I am persuaded that life, as it is known to us,*
*is a direct result of the asymmetry of the universe or of its indirect conse-*
*quences. The universe is asymmetric."* — Louis Pasteur

Multi-agent systems (MAS) often include a combinatorial problem which is distributed among the agents. Examples of multi-agent combinatorial problems are the Meetings Scheduling problem (Modi & Veloso, 2004; Gershman, Grubshtein, Meisels, Rokach, & Zivan, 2008), Sensor nets (Zhang, Xing, Wang, & Wittenburg, 2005; Zivan, Glinton, & Sycara, 2009), and Vehicle Routing (Léauté & Faltings, 2011). The natural representation of such problems in terms of agent-owned variables and in terms of agent-specified values for combinations of assignments (whether costs or utilities), has encouraged the study of Distributed Constraint Optimization Problems (DCOPs). DCOPs are a powerful framework for formulating and solving MAS combinatorial problems. In the last decade algorithmic search techniques for DCOPs were intensively studied (Yeoh, Felner, & Koenig, 2010; Petcu & Faltings, 2006; Mailler & Lesser, 2004; Gershman, Meisels, & Zivan, 2009). Since DCOPs are NP-hard, many recent studies consider incomplete algorithms (Maheswaran, Pearce, & Tambe, 2006; Zhang et al., 2005; Pearce & Tambe, 2007; Zivan, 2008; Rogers, Farinelli, Stranders, & Jennings, 2011).

The strong relation between MAS and DCOPs was identified and discussed in previous studies (e.g., Maheswaran et al., 2006; Chapman, Rogers, & Jennings, 2008). Chapman et al. (2008) examine the analogy between the DCOP formulation and a class of games known as *"Potential Games"*. The importance of this analogy lies in the fact that every finite potential game possesses at least one pure strategy Nash Equilibrium (NE) (Monderer & Shapley, 1996).

In the world of game theory, a pure strategy NE is a stable profile of actions corresponding to the set of all participants, in which any unilateral change of action by a single participant will not yield a better *personal* gain for the participant. In the DCOP formulation, the above definition coincides with special solutions known as local optima (minima or maxima) (Yokoo, 2000; Zhang et al., 2005). These solutions are sets of assignments to variables made by all agents, in which a single change of assignment by an agent will only reduce the *global* gain.

The source of this correspondence between NEs and local optima stems from the constraint structure of DCOPs. Each constraint $C$ over variables of $k$ agents is defined as a mapping from the domains of the variables to a single (positive) real value:

$$C : D_{i_1} \times D_{i_2} \times \cdots D_{i_k} \to \mathbb{R}_+$$

The above definition of a constraint implies that the cost (gain) of a constraint is the same for all participating agents. When an agent lowers its cost or gain from a constraint, all of its constrained peers share a similar decrement in cost from that constraint. Thus, it is clear that a change of an assignment can reduce personal gains if and only if it reduces global gains as well.

In many real life situations constrained agents value differently the results of decisions on constrained issues even if they consider the same constraints. In fact, this is the natural scenario in a typical MAS situation as in the following examples. In the meeting scheduling problem, agents which attend the same meeting may derive different utilities from it. Moreover, their preferences and constraints regarding its time and location are expected to be different. Another example for a distributed application which is asymmetric is the smart grid (Ramchurn, Vytelingum, Rogers, & Jennings, 2011), where the cost users pay for electricity may be higher in heavy load hours, yet the increase of price is endured with respect to the agents' use and is not evenly spread among users. Supply chain management (Burke, Brown, Dogru, & Lowe, 2007) among multiple consumers might also include asymmetric constraints. Different consumers can have different levels of urgency regarding the time of supply, therefore the latency costs they endure are different.

The above observation calls for a generalization of the standard DCOP model (Modi, Shen, Tambe, & Yokoo, 2005; Meisels, 2007). Such a generalized model will enable the representation of asymmetric gains for agents involved in a constraint. As a result, it will be applicable to asymmetric MAS scenarios.

Former studies proposed to capture asymmetric gains among constrained agents by introducing additional variables for each agent. The additional variables are duplicates of the variables of constrained agents. Each agent holds duplications for every variable its own variables are constrained with. By imposing hard equality constraints between variables and their duplications the model allows each agent to account for its own constraints (Maheswaran, Tambe, Bowring, Pearce, & Varakantham, 2004b; Petcu, 2007). The complete scheme of duplicating all variables of constrained agents and of using rigid constraints to enforce equality of assignments with other agents was termed Private Events as Variables (PEAV).

When considering complete search, the PEAV formulation indeed offers a solvable representation to an asymmetric DCOP that allows the use of algorithms designed to solve symmetric DCOPs. The main consequence of using this model is the increment in the problem size, which (as is demonstrated in the experimental evaluation of the present paper) for an NP-hard problem such as DCOP, has a devastating effect on performance.

The situation becomes more complicated when considering incomplete methods. Specifically, PEAV does not enable the use of standard local search algorithms for solving asymmetric problems. The present paper demonstrates that every allocation which satisfies the hard equality constraints in PEAV is a local optimum which cannot be escaped by local search algorithms.

The present study proposes Asymmetric DCOPs (ADCOPs), a model for representing asymmetric combinatorial multi-agent problems. ADCOPs naturally accommodate constraints where the participating agents have different gains or costs. It allows each agent to hold its own evaluation of different outcomes with respect to each constraint it is involved in.

The shortcomings of existing DCOP algorithms for solving problems represented by the proposed model triggers an intensive algorithmic study:

1. for complete search we propose algorithms that can solve an asymmetric problem without the need to expand the number of variables of the problem as in the PEAV model. The advantages in performance of the proposed algorithms over state-of-the-art complete algorithms that use the PEAV representation are demonstrated empirically.

2. for incomplete (local) search we propose algorithms that are able to converge to high quality solutions when solving asymmetric problems, in contrast to existing DCOP local search algorithms. A proof that guarantees this convergence is provided. The proposed algorithms require some exchange of the problem's information among agents. Thus, the algorithms are evaluated not only in terms of solution quality, but in terms of privacy as well. The empirical results of the present paper demonstrate that the privacy loss of the proposed algorithms is minor.

The rest of this paper is organized as follows. The proposed Asymmetric DCOP model is presented in Section 2 along with several alternatives for representing asymmetric constraints within the standard DCOP model. Section 3 introduces new complete search algorithms that are designated to efficiently and correctly solve asymmetric problems. Section 4 focuses on local search. It demonstrates the incompatibility of existing local search algorithms for solving asymmetric problems and proposes several novel asymmetric local search algorithms. Section 5 includes an extensive experimental evaluation of both complete search algorithms and local search algorithms. The results of the experimental evaluation are quite conclusive. The paper is summarized and conclusions are drawn in Section 6.

## 2. Asymmetric Distributed Constraint Optimization

First, the standard DCOP model is presented followed by its proposed generalization that captures asymmetric constraints. Next, some alternatives for representing asymmetric constraints by the standard DCOP model are described.

### 2.1 Distributed Constraint Optimization Problems (DCOPs)

A *DCOP* is a tuple $\langle \mathcal{A}, \mathcal{X}, \mathcal{D}, \mathcal{R} \rangle$. $\mathcal{A}$ is a finite set of agents $A_1, A_2, ..., A_n$. $\mathcal{X}$ is a finite set of variables $X_1, X_2, ..., X_m$. Each variable is held by a single agent (an agent may hold more than one variable). $\mathcal{D}$ is a set of domains $D_1, D_2, ..., D_m$. Each domain $D_i$ contains a finite set of values which can be assigned to variable $X_i$. $\mathcal{R}$ is a set of relations (constraints). Each constraint $C \in \mathcal{R}$ defines a non-negative *cost* for every possible value combination of a set of variables, and is of the form:

$$C : D_{i_1} \times D_{i_2} \times \ldots \times D_{i_k} \to \mathbb{R}_+ \tag{1}$$

A *binary constraint* refers to exactly two variables and is of the form $C_{ij} : D_i \times D_j \to \mathbb{R}_+$. A *binary DCOP* is a DCOP in which all constraints are binary. A *value assignment* is a pair including a variable, and a value from that variable's domain. A *partial assignment* (PA) is a set of value assignments, in which each variable appears at most once. A *full assignment* or a *solution* is a partial assignment that includes all the variables ($vars(PA) = \mathcal{X}$). An *optimal solution* is a full assignment of aggregated minimal cost.

In maximization problems, for each constraint we have utilities instead of costs and a solution is a full assignment of maximal aggregated utility. In the rest of this paper, unless stated differently, the problems discussed are minimization problems.

### 2.2 Asymmetric DCOPs (ADCOPs)

ADCOPs generalize DCOPs in the following manner: instead of assuming equal payoffs for constrained agents, the ADCOP constraint explicitly defines the exact payoff for each participant. That is, domain values are mapped to a tuple of costs, one for each constrained agent.

More formally, an ADCOP is defined by the following tuple $\langle \mathcal{A}, \mathcal{X}, \mathcal{D}, \mathcal{R} \rangle$, where $\mathcal{A}$, $\mathcal{X}$, and $\mathcal{D}$ are defined in exactly the same manner as in DCOPs. Each constraint $C \in \mathcal{R}$ of an asymmetric DCOP defines a set of non-negative *costs* for every possible value combination of a set of variables, and takes the following form:

$$C : D_{i_1} \times D_{i_2} \times \cdots D_{i_k} \to \mathbb{R}_+^k \qquad (2)$$

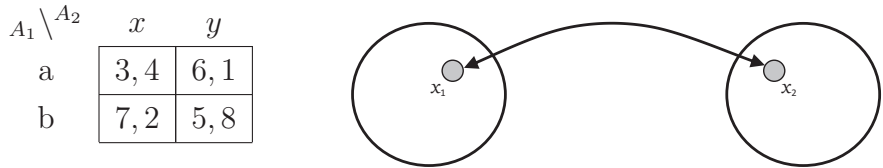

Figure 1: A two agents interaction. The left-hand side presents the agents' costs incurred by the interaction – Agent $A_1$'s possible value assignments are either $a$ or $b$ and the costs they endure are depicted in the left value in each cell. Agent $A_2$'s possible value assignments are $x$ or $y$ (costs on the right side). The right-hand side of the figure provides a graphical presentation the resulting ADCOP network.

This definition of an asymmetric constraint is natural for general MAS problems, and requires little manipulation when formulating these as ADCOPs. It applies when each agent holds exactly one variable among the $k$ variables involved in the constraint $C$. We note that when several variables involved in a constraint are held by the same agent, the length of the vector of costs representing the constraint is equal to the number of agents involved and not the number of variables.

Consider the simple example of two interacting agents presented in Figure 1. In this problem each agent pays either the left- or right-hand side of the values depicted in the bi-matrix of Figure 1. Agent $A_1$ controls variable $x_1$ and may assign either $a$ or $b$ to it and Agent $A_2$ controls variable $x_2$ and may assign the values $x$ or $y$. The constraint between the two interacting agents maps assignment pairs to *cost pairs*. For example, if agent $A_1$ assigns the value $b$ and agent $A_2$ assigns the value $x$, $A_1$'s cost will be 7 and $A_2$'s cost will be 2. We often refer to the cost pairs as two *sides* of a constraint. The right-hand side of the figure illustrates the ADCOP formulation of this problem in terms of agents and their variables (where the single constraint is presented in full in the bi-matrix on the left).

It should be noted that although the above ADCOP model bears great resemblance to the graphical games model (Kearns, Littman, & Singh, 2001) these two models are fundamentally different. While game-theoretic agents are self-interested entities, ADCOP agents are cooperative by nature and always follow the protocol (algorithm) – even at the risk of personal degradation of gain.

## 2.3 Alternatives for Representing Asymmetric Constraints

Extending the Distributed Constraint Reasoning (DCR) (Meisels, 2007) paradigm to encompass asymmetric payoffs is of great interest when considering real world problems. Typical problems must often take into account the individual state of each agent (remaining battery life, user preferences, etc) which rarely coincide. Several alternatives for representing asymmetric constraints by DCOPs were discussed in former papers.

### 2.3.1 Disclosure of Constraint Payoffs

The simplest way to solve MAS problems with asymmetric payoffs by DCOPs is through the disclosure of constraint payoffs. That is, aggregate values of all agents taking a joint action. However, constraint disclosure reveals private information (preferences) (Greenstadt, Pearce, & Tambe, 2006; Maheswaran, Pearce, Varakantham, Bowring, & Tambe, 2005; Yokoo, K.Suzuki, & Hirayama, 2002) and many times needs to be avoided.

### 2.3.2 Use of Unary Constraints

A possible technique for representing preferences and affecting different payoffs is through the introduction of unary constraints. Constraints are added to each variable participating in a constraint and the additional costs generate asymmetry.

**Proposition 1** *There exists an asymmetric DCOP that cannot be expressed by a symmetric DCOP with the addition of unary constraints of the form $U_{A_i}(\cdot) = \alpha$.*

**Proof:** Consider for example the interaction depicted in Figure 1. The most general solution is to add an unary constraint for each variable held by the agents. The cost of the binary constraint, $B(\cdot, \cdot)$, and unary constraint of each variable $U_A(\cdot)$, $U_B(\cdot)$ must be consistent with the cost incurred on each agent, and can be described by the following set of linear equations:

$$U_{A_1}(a) + B(a, x) = 3$$

$$U_{A_1}(a) + B(a, y) = 6$$

$$\vdots$$

$$U_{A_2}(y) + B(b, y) = 8$$

This set of equations that is derived from the problem of Figure 1 has no solution. $\square$

**Corollary 2** *Symmetric DCOPs with the addition of unary constraints are strictly less expressive than asymmetric DCOPs.*

**Proof:** It remains to show that every problem with unary and symmetric binary constraints can be represented as an asymmetric DCOP. Unary constraints are an integral part of the asymmetric DCOP model, while symmetric binary constraints are inherently less expressive than asymmetric ones. Consequently, every symmetric DCOP (with added unary constraints) can be represented as an asymmetric DCOP. $\square$

The unary constraints approach is thus shown to be insufficient for representing asymmetric constraints. More precisely, this approach fails to properly capture cases in which the personal valuation of a state by an agent is dependent upon assignments by other agents.

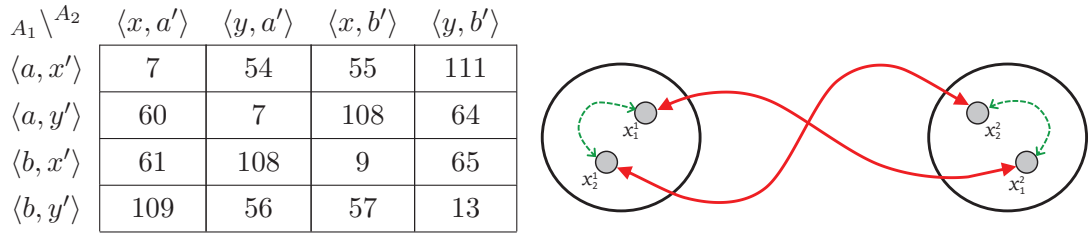

| $A_1 \backslash^{A_2}$ | $\langle x, a' \rangle$ | $\langle y, a' \rangle$ | $\langle x, b' \rangle$ | $\langle y, b' \rangle$ |
|---|---|---|---|---|
| $\langle a, x' \rangle$ | 7 | 54 | 55 | 111 |
| $\langle a, y' \rangle$ | 60 | 7 | 108 | 64 |
| $\langle b, x' \rangle$ | 61 | 108 | 9 | 65 |
| $\langle b, y' \rangle$ | 109 | 56 | 57 | 13 |

Figure 2: The same interaction of Figure 1 formulated as a PEAV-DCOP. The left-hand side presents the value of each possible end state. The right-hand side provides a graphical representation of the resulting PEAV-DCOP network.

### 2.3.3 Private Events As Variables (PEAV)

The PEAV model (Maheswaran et al., 2004b) successfully captures asymmetric payoffs within standard DCOPs. The incurred cost of the PEAV model on an agent involves one mirror variable per each of its neighbors in the constraint network (i.e., for each constrained variable held by another agent). Consistency with the neighbors' state variables is imposed by a set of hard equality constraints. One way to represent hard constraints is to assign a cost that is calculated for each specific problem (Maheswaran et al., 2004b).

The resulting representation of an asymmetric MAS problem in a PEAV-DCOP is much larger in terms of variables and constraints than an ADCOP. Figure 2 describes the same interaction of Figure 1 formulated according to PEAV (note that this is a minimization problem). In this example $x_1^1$ and $x_2^2$ are the original variables and $x_2^1$ and $x_1^2$ are the mirror variables generated by the PEAV formulation. The upper bound value used as a hard constraint in this example is 50.

The cost of each end state is depicted in the table on the left-hand side of the figure. For example, the top-left cell represents the assignments $x_1^1 = a$, $x_2^2 = x$, and consistent values of the mirror variables. Consequently, the cost (7) is exactly the sum of costs of both agents in the original MAS problem (Figure 1).

The PEAV representation of this problem must accommodate the mirror variables, and combinations of values previously not considered must also be taken into account. For example, the top-right cell represents the assignments $x_1^1 = a$, $x_2^2 = y$, but here the mirror variables are not consistent with the original variables ($x_2^1 = x$, $x_1^2 = b$). The cost (111) includes the sum of costs of both agents in the original MAS problem, where each agent considers the value of the mirror variable as the real value of the other agent (3 and 8). To this value, two upper bound values (50 each) were added to express the inconsistency of the mirror variables. Consequently, the PEAV matrix is $4 \times 4$, including 16 payoff values – a quadratic increase in the size of the search space.

## 3. Asymmetric Distributed Complete Search

Although DCOPs are NP-hard, a large number of former DCOP research was dedicated to complete search algorithms (Modi et al., 2005; Yeoh et al., 2010; Petcu & Faltings, 2006;

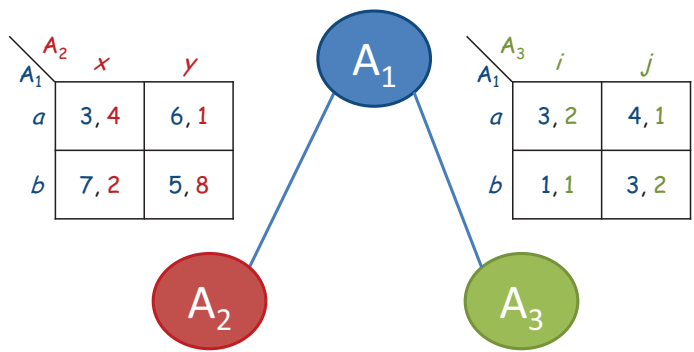

Figure 3: Example problem.

 Unfortunately, most existing complete DCOP algorithms cannot find the optimal solution when solving ADCOPs. We elaborate on this issue at the beginning of this section and then introduce several novel complete ADCOP algorithms.

## 3.1 Complete Search with Asymmetric Constraints

One may attempt to solve ADCOPs by simply using existing complete DCOP algorithms. The following scenario demonstrates the shortcomings of such an attempt. Consider the SyncBB algorithm (Hirayama & Yokoo, 1997). SyncBB was chosen for its simplicity, however, this demonstration is relevant for most existing DCOP algorithms. In SyncBB a partial assignment is generated sequentially on a Current Partial Assignment (CPA) message (cf., Meisels, 2007). When an agent receives the CPA, it assigns its variable so that the overall cost of the partial assignment is minimal. Running a Branch & Bound algorithm implies that whenever the cost of a partial assignment becomes larger than the upper bound, the agent which holds the CPA backtracks to the agent before it.

For ADCOPs the above standard process is not correct. The full cost of a partial assignment must include the constraints held by all the agents which participate in it. In other words, once an agent $A_i$ assigns its variables it can only compute the cost of the CPA based on its own evaluation, while other agents with higher priority (i.e., ordered *before it*) are holding constraints between the agent's newly assigned variable and their own. Consider the example problem in Figure 3. When attempting to solve this problem using SyncBB, only the parts of the constraints held by the lower priority agents in each constraint are evaluated. Thus, the resulting solution is $\langle A_1 = a, A_2 = y, A_3 = j \rangle$ with a cost of 2. The constraints held by higher priority agents ($A_1$ in this example) are not evaluated. The real cost of this solution when considering both sides of the constraints is 12. The actual solution is $\langle A_1 = b, A_2 = x, A_3 = i \rangle$ with a cost of 11.

Following the above example, it is clear that most existing complete DCOP algorithms are no longer correct when the ADCOP model is used. A complete ADCOP algorithm must allow all agents participating in a constraint to evaluate the related assignments. An exception may be the OptAPO algorithm (Mailler & Lesser, 2006), since it does not perform distributed search in order to resolve conflicts (Grinshpoun & Meisels, 2008). Thus, when (mediator) agents perform search they are able to generate a centralized ("symmetric") problem and solve it.

## 3.2 ADCOP Complete Search Algorithms

Solving an asymmetric problem requires that both parts of each binary constraint are evaluated and aggregated in the assignment's cost. Similarly to the DCSP case (Brito, Meisels, Meseguer, & Zivan, 2009) considering both sides of the constraints can be generally performed in two ways. One strategy is to solve the problem in two phases. In the first phase, a full assignment is reached while considering only one side of each constraint – this phase can be performed using a symmetric DCOP algorithm. In the second phase, the full cost of the assignment is verified by checking the complementary part of all the involved constraints. After the second phase is complete and new bounds were set, the first phase is resumed in a search for a better solution until the entire search space is covered. Another strategy is to systematically check both sides of the constraints before reaching a full assignment, forming a one-phase strategy. The present paper refers to checking the reversed-order part of the constraints as *back-checking*. Back-checking can be performed either synchronously or asynchronously.

We next present several innovative complete algorithms designed for ADCOPs. We begin by introducing two asymmetric versions for the SyncBB algorithm. The first version, SyncABB-2ph, follows the two-phase strategy, while the other version follows the one-phase strategy. We also present the Asynchronous Two-Way Bounding algorithm (ATWB), which is an asymmetric version for the AFB algorithm (Gershman et al., 2009). ATWB also follows the one-phase strategy and naturally performs asynchronous back-checking.

### 3.2.1 Synchronous Asymmetric Branch & Bound – 2-phase (SyncABB-2ph)

The SyncABB-2ph algorithm is a combination of the SyncBB algorithm with the two-phase strategy. In phase I the algorithm works exactly as SyncBB, where each agent counts the costs of its constraints with lower-indexed agents. Phase I is completed when a full assignment is reached. In standard (symmetric) SyncBB operating on a symmetric DCOP, this means that a new best solution and a new bound has been found. For ADCOPs back-checking is needed in order to verify that the reversed order parts of the constraints do not increment the cost beyond the bound.

---

**Algorithm 1** SyncABB-2ph: phase II

**when received ⟨CPA_BACK_MSG, $CPA$, $cost$⟩ do**
  1: $f \leftarrow$ cost of constraints with higher-indexed agents $(A_{i+1}...A_n)$
  2: **if** $cost + f \geq B$ **do**
  3:   send ⟨CPA_MSG, $CPA$⟩ to $A_n$
  4: **else if** $A_i \neq A_1$ **do**
  5:   send ⟨CPA_BACK_MSG, $CPA$, $cost + f$⟩ to $A_{i-1}$
  6: **else**
  7:   $B \leftarrow cost + f$
  8:   broadcast ⟨NEW_SOLUTION, $CPA$, $B$⟩
  9:   send ⟨CPA_MSG, $CPA$⟩ to $A_n$

---

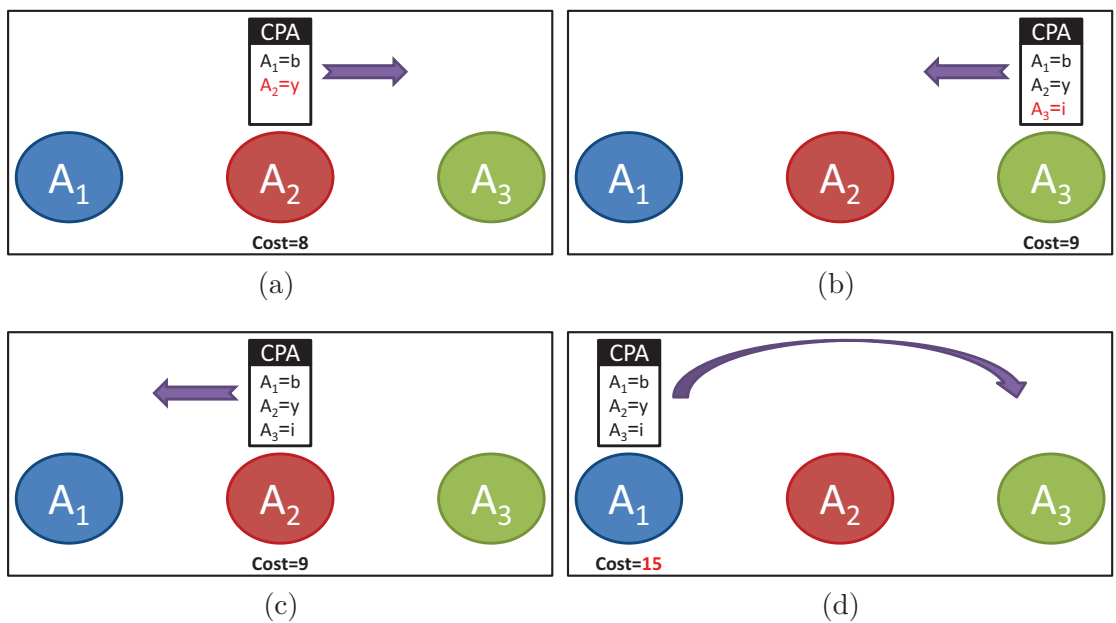

Figure 4: SyncABB-2ph example – limited pruning.

In phase II the last agent $(A_n)$ sends to its preceding agent $(A_{n-1})$ a **CPA_BACK_MSG** message. This message includes the CPA and its one-side cost that was gathered in phase I. Each agent that receives the **CPA_BACK_MSG** message (Algorithm 1) performs back-checking by computing the cost $f$ of its constraints with upper-indexed agents (line 1). If the addition of $f$ to the cost of the CPA (the lower bound) reaches the bound $B$ (line 2), there is no point to continue the back-checking phase since a value assignment must be replaced. To ensure the completeness of the algorithm, regardless of which agent identified that the cost exceeded the global bound (the lowest cost for a full assignment found so far), the CPA must be returned to the last agent $A_n$ (line 3). If the global bound has not been reached, the back-checking continues to the preceding agent (lines 4-5) until it reaches $A_1$. When the total cost of the CPA is below the existing global bound, agent $A_1$ updates the bound $B$ (line 7) and informs all the agents of the new best solution and new bound (line 8). Next, phase I is resumed by sending the CPA back to the last agent (line 9).

During the solving process parts of the search space are expected to be pruned. In fact, the amount of pruned search space faithfully reflects the effectiveness of a Branch & Bound algorithm. Lines 2 and 3 in Algorithm 1 state that when the global bound is reached during the back-checking phase, the CPA is returned to the last agent. Returning the CPA to the last agent means no pruning. Consequently, during the run of the SyncABB-2ph algorithm pruning can only be achieved in the first phase. This limits the pruning that can be performed by the algorithm since in the first phase only a subset of the constraints are considered. This shortcoming of the algorithm is illustrated in Figure 4. When the value of $A_2$ is assigned (stage a), the CPA is $\langle A_1 = b, A_2 = y \rangle$. A global look at the CPA (that considers both sides of the constraint) reveals that its cost is 13, which is higher than the current bound (11 at this stage of the search). Nevertheless, at the first phase of the algorithm only one side of the constraint is evaluated, resulting in a cost of 8. Thus, the CPA advances and agent $A_3$ also assigns a value (stage b). The bound is only passed during the second phase (stage d), and so the CPA is returned back to agent $A_3$, which in turn will assign the value $A_3 = j$.

The demonstrated pruning problem may very well result in poor performance of the SyncABB-2ph algorithm. Indeed, our experimental evaluation supported this conjecture, and so the two-phase approach was abandoned.

### 3.2.2 Synchronous Asymmetric Branch & Bound – 1-phase (SyncABB)

The SyncABB algorithm is a combination of the SyncBB algorithm with the one-phase strategy. After each step of the algorithm, when an agent adds an assignment to the CPA and updates the cost with one side of the constraint, the CPA is sent back to the agents that have already assigned their variables to update the lower bound by adding the costs of all backwards directed constraints (back-checking). This is done by replacing the **CPA_MSG** message sent after each value assignment to the next agent (as in SyncBB and SyncABB-2ph) with a **CPA_BACK_MSG** message to the preceding agent.

---

**Algorithm 2** SyncABB: back-checking

**when received $\langle$CPA_BACK_MSG, $CPA, cost\rangle$ do**
1:  $j \leftarrow CPA.lastId$
2:  $f \leftarrow$ cost of constraint with agent $A_j$
3:  **if** $cost + f \geq B$ **do**
4:     send $\langle$**CPA_MSG**, $CPA\rangle$ to $A_j$
5:  **else if** $A_i \neq A_1$ **do**
6:     send $\langle$**CPA_BACK_MSG**, $CPA, cost + f\rangle$ to $A_{i-1}$
7:  **else if** $A_j = A_n$ **do**
8:     $B \leftarrow cost + f$
9:     broadcast $\langle$**NEW_SOLUTION**, $CPA, B\rangle$
10:    send $\langle$**CPA_MSG**, $CPA\rangle$ to $A_n$
11:  **else**
12:    $CPA.cost \leftarrow cost + f$
13:    send $\langle$**CPA_MSG**, $CPA\rangle$ to $A_{j+1}$

---

The handling of a **CPA_BACK_MSG** in SyncABB is presented in Algorithm 2. First, it is important to know the identity $j$ of the initiator of the back-checking (in SyncABB-2ph it was always $n$). Consequently, agent $A_n$ (in SyncABB-2ph) is now replaced by $A_j$ (lines 2,4). Additionally, when the back-checking is complete (reaches $A_1$) and $A_j$ is not the last agent (line 11), the algorithm simply sends the CPA to the next assigning agent $A_{j+1}$ (line 13).

Figure 5 illustrates how the CPA is moved between the agents in the SyncABB algorithm. The given example shows the run of the algorithm at the beginning of the search process for the problem depicted in Figure 3.

The main motivation for this one-phase version is that when the global bound is reached, the CPA can be returned to the initiator of the back-checking (line 4), which in many cases will not be $A_n$. This can lead to effective pruning of the search space in comparison to the two-phase strategy. This behavior of the algorithm is illustrated in Figure 6. When the value of $A_2$ is assigned (stage a), the CPA includes $\langle A_1 = b, A_2 = y \rangle$. After the back-checking is performed (stage b) the cost becomes 13 and so the bound (11 at this stage of the search) is reached. The CPA is passed at that point back to agent $A_2$. Consequently, some of the search space is pruned (complete assignments $\langle A_1 = b, A_2 = y, A_3 = i \rangle$ and $\langle A_1 = b, A_2 = y, A_3 = j \rangle$).

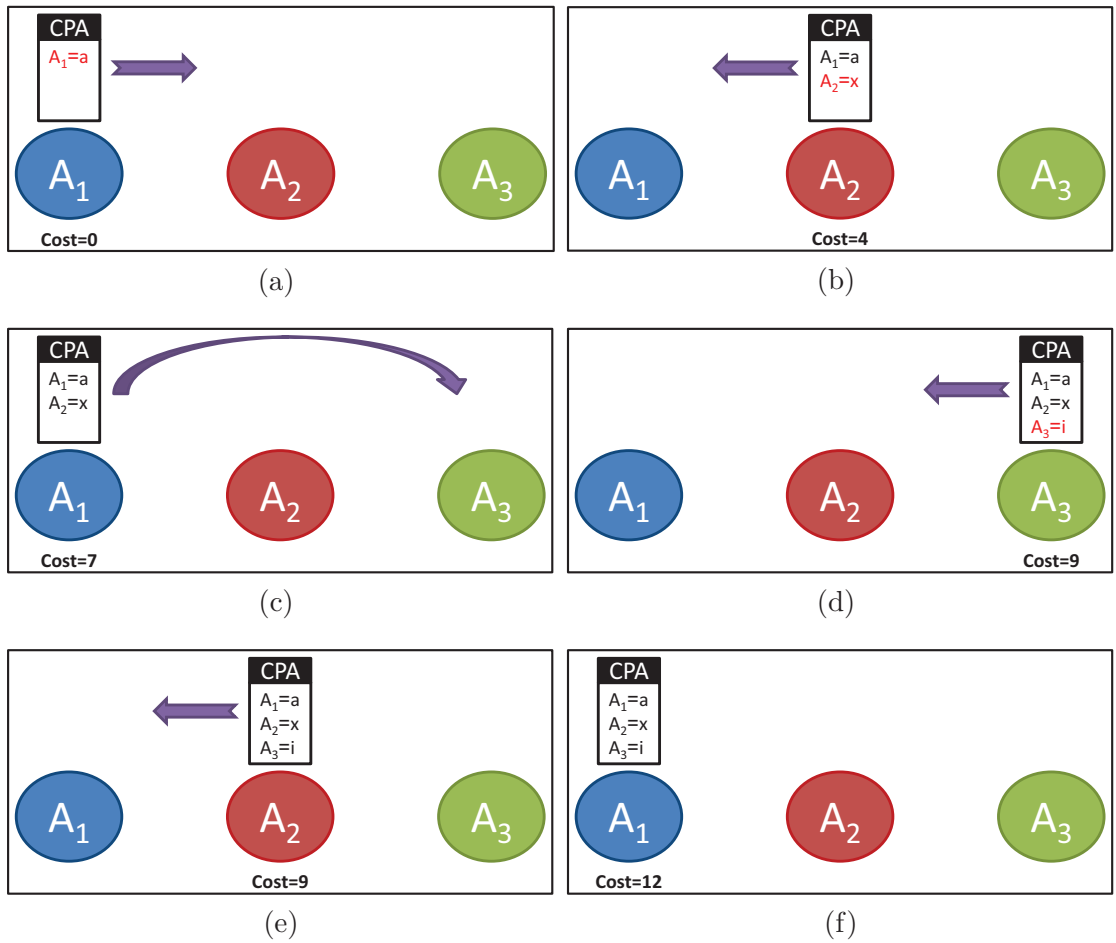

Figure 5: SyncABB-1ph example.

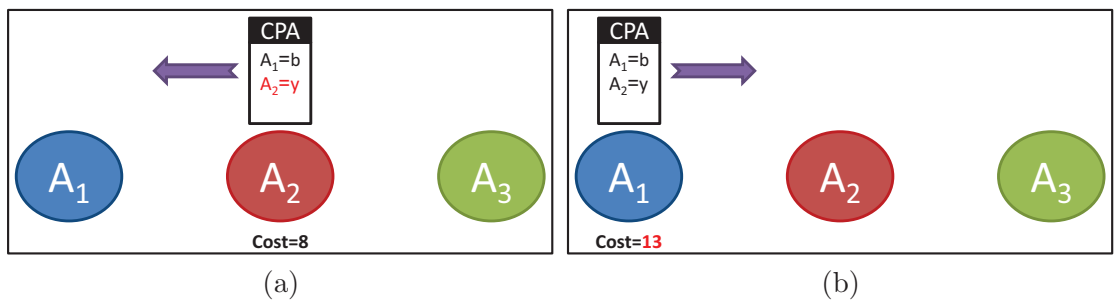

Figure 6: SyncABB-1ph example – effective pruning.

**Proposition 3** *SyncABB is sound and complete.*

**Proof:** The soundness of SyncABB is immediate since a CPA is sent forward by an assigning agent only after the new assignment was evaluated by all agents whose assignment is included in the CPA. In other words, a CPA is sent forward only after all costs generated by the addition of the new assignment were added to the lower bound on the CPA (Algorithm 2, lines 12-13). This includes the complete assignment in a reported solution (lines 8-10).

The completeness of SyncABB follows from the exhaustive search structure. Only partial assignments whose cost exceeds the global bound are not extended and therefore the existence of a solution with a lower cost than the solution found by the algorithm is ruled out.

Termination also follows from the exhaustive structure of the Branch & Bound algorithm in which no partial assignment can be explored twice. □

### 3.2.3 Asymmetric Two-Way Bounding (ATWB)

To achieve a larger degree of asynchronicity, one can build upon an existing and efficient asynchronous DCOP algorithm, AFB (Gershman et al., 2009). AFB was found to perform faster than competing complete algorithms such as DPOP and ADOPT. In the AFB algorithm agents assign a CPA sequentially as in SyncBB, but following each assignment the assigning agent triggers asynchronous checks of bounds by sending copies of the CPA via **BOUND_CPA** messages to agents which have not yet assigned their variables. The agents that receive copies of the CPA calculate the lower bound on the cost of the constraints they hold with the assignments on the CPA. The lower bound is sent back to the assigning agent via an **ESTIMATE** message. The assigning agent aggregates the bounds in the **ESTIMATE** messages it receives into an updated lower bound and if it exceeds the current upper bound, the agent initiates a backtrack.

The AFB algorithm can be adjusted to accommodate both forward bounding and backward bounding and thus, can be adjusted for solving ADCOPs. In other words, instead of sending each assigned CPA back to the first assigning agent sequentially as in SyncABB, copies of the CPA can be sent backwards to agents whose assignments are included in the CPA. Agents that receive a copy of the CPA compute their estimate and send it back (forward) to the assigning agent just as in standard AFB. We refer to this version as the Asynchronous Two-Way Bounding algorithm (ATWB).

The ATWB algorithm follows the pseudo-code of AFB (Gershman et al., 2009) with several modifications. **BOUND_CPA** messages are now sent *both forward and backwards* whenever a value is assigned (procedure **assign_CPA**). Additionally, the last agent $A_n$ cannot declare a new solution until it receives all the estimates from the backward bounding. Thus, the handling of **ESTIMATE** messages must be revised and its new version is given in Algorithm 3.

When an estimate message is received the agent checks whether the new estimate reaches the global bound (line 2). If this is the case, then a new value is assigned by the current agent (line 3). In case this is the last agent and all the backward estimates have arrived (line 4), the agent can declare a new solution (lines 5-6) and assign a new value (line 7).

---

**Algorithm 3** ATWB – receive estimate

---

**when received** (**ESTIMATE**, *estimate*)
1: save *estimate*
2: **if** ($CPA.cost$ + all saved estimates)$\geq B$ **do**
3:    assign_CPA()
4: **else if** $CPA$ is a full assignment **and** all estimates arrived **do**
5:    $B \leftarrow CPA.cost$ + all saved estimates
6:    broadcast (**NEW_SOLUTION**, $CPA$, $B$)
7:    assign_CPA()

---

In the case of forward bounding, an agent still does not know which assignment it will select and therefore its estimate is a lower bound on the cost of the constraints considering all values in its variable's domain. In contrast, when an agent $A_i$ receives a **BOUND_CPA** message from an agent $A_j$ ordered after it (backward bounding), it can accurately compute the cost of the constraints of its own assignment with all other assignments on the CPA. A precomputed $h_2(v, j)$ function (per value $v$, per agent $A_j$ which holds the current CPA) that gives a lower bound on the cost of constraints with agents that are after $A_j$ in the order ($\forall A_k | k > j$) can be added to the estimation computation. The $h_2(v, j)$ function can also be used in forward bounding as a lower bound for the back-checking of value $v$ with all the agents between $A_i$ and $A_j$ ($\forall A_k | j < k < i$). The $h_2$ function is additive, since it refers to the back-checking of yet unassigned variables.

**Proposition 4** *ATWB is sound and complete.*

**Proof:** The soundness of ATWB is established by the fact that a new solution is stored and later reported only after all estimates arrive at the last agent (Algorithm 3, lines 4-6). At that point, all constraints have been evaluated by all involved agents. Note that the estimates received by the last agent include all (backward) constraints. The possibility that some estimate was not yet received by some agent (due to delay of messages) does not compromise the algorithm's soundness. In case the delayed estimate would have triggered a need to backtrack, the estimate sent by the same agent to the last agent would be at least as high and therefore would trigger a backtrack as well.

Similarly to SyncABB, the completeness and termination of ATWB follow from the exhaustive structure of the Branch & Bound algorithm. □

## 4. Asymmetric Distributed Local Search

Distributed *local search* techniques for solving DCOPs have gained popularity in recent years. Although local search algorithms are inherently incomplete, i.e. they do not guarantee to report the optimal solution, they offer a practical solution for significantly larger problems. Adding asymmetry to the problems makes a complete solving process even more difficult, a fact that enhances the suitability of local search for solving ADCOPs.

We next present an overview of standard DCOP local search algorithms followed by a discussion of their applicability to the asymmetric case. We then introduce several novel local search algorithms designed for solving ADCOPs.

## 4.1 Local Search

The general design of local search algorithms for DCOPs is synchronous. In each step of the algorithm an agent sends its assignment to all its neighbors in the constraint network and receives the assignments of all its neighbors. We hereby present in detail two leading algorithms that apply to this general framework – the Distributed Stochastic Algorithm (DSA) (Zhang et al., 2005) and the Max Gain Message (MGM) algorithm (Maheswaran, Pearce, & Tambe, 2004a).[1]

In the initial step of the DSA algorithm agents randomly pick some value assignment for their variable. Next, agents perform a sequence of steps until some termination condition is met. In each step, each agent sends its value assignment to its neighbors in the constraints graph and receives the assignments of its neighbors. The present paper follows the general definition of a DCOP which does not include a synchronization mechanism. If such a mechanism exists, agents in DSA can send value messages only in steps in which they change their value assignments. After collecting the assignments of all its neighbors, each agent decides whether to keep its value assignment or to change it, by using a stochastic strategy (see Zhang et al., 2005 for details on the possible strategies and the difference in the resulting performance). A sketch of DSA is presented in Algorithm 4.

---
**Algorithm 4** Standard DSA
---
1: $value \leftarrow$ ChooseRandomValue()
2: **while** no termination condition is met **do**
3:     send $value$ to neighbors
4:     collect neighbors' values
5:     **if** ReplacementDecision() **do**
6:         select and assign the next value
---

The MGM algorithm is a strapped down version of the DBA algorithm (Yokoo, 2000; Zhang et al., 2005). In every synchronous step, each agent sends its current value assignment to its neighbors and collects their current value assignments. After receiving the assignments of all its neighbors, the agent computes the maximal improvement (i.e., reduction in cost) to its local state that can be achieved by replacing its assignment and sends this proposed reduction to its neighbors. After collecting the proposed reductions from its neighbors, each agent changes its assignment only if its proposed reduction is greater than the reductions proposed by all of its neighbors. In more advanced versions of MGM, agents group together in order to propose a common improvement and thus avoid local minima to which a smaller group would have converged. Algorithm 5 includes a sketch of the standard MGM algorithm. After selecting a random value for its variable (line 1), the agent enters a loop where each iteration is a step of the algorithm. After sending its value assignment to its neighbors and collecting their value assignments (lines 3-4), the agent calculates its best weight reduction and sends it to its neighbors (lines 5-6). After receiving the possible weight reductions of all of its neighbors the agent decides whether to replace its assignment and upon a positive decision reassigns its variable (lines 7-10).

---

1. Our description considers an improvement to be a decrease in the number of violated constraints (as in Max-CSPs).

---

**Algorithm 5** Standard MGM

---
1:  $value \leftarrow$ ChooseRandomValue()
2:  **while** no termination condition is met **do**
3:      send $value$ to neighbors
4:      collect neighbors' values
5:      $LR \leftarrow$ BestPossibleLocalReduction()
6:      send $LR$ to neighbors
7:      collect neighbors' $LRs$
8:      **if** $LR > 0$ **do**
9:          **if** $LR > LRs$ of neighbors (ties broken using indexes) **do**
10:             $value \leftarrow$ the value that gives $LR$

---

A different incomplete approach for solving DCOPs is implemented in the Max-Sum algorithm (Farinelli, Rogers, Petcu, & Jennings, 2008). The Max-Sum algorithm operates on a *factor graph* which is a bipartite graph in which the nodes represent variables and constraints.[2] Each node representing a variable of the original DCOP is connected to all function-nodes that represent constraints which it is involved in. Similarly, a function-node is connected to all variable-nodes that represent variables in the original DCOP which are included in the constraint it represents. Agents in Max-Sum perform the roles of different nodes in the factor graph. We will assume that each agent takes the role of the variable-nodes which represent her own variables and for each function-node, one of the agents who's variable is involved in the constraint it represents, performs its role. Variable-nodes and function-nodes are considered as "agents" in Max-Sum, i.e., they can send messages, read messages, and perform computation.

The content of messages sent by function-nodes is different than the content of messages sent by variable-nodes. A message sent from a variable-node to a function-node includes for each of its possible value assignments, the sum of costs/utilities for this value it received from all other function neighbors. A message sent from a function-node to a variable-node includes for each possible value assignment of the variable the best (minimal in a minimization problem, maximal in a maximization problem) cost/utility that can be achieved from any combination of assignments to the variables involved in the function not including costs/utilities reported by the destination variable. At the end of the run each variable selects the value assignment that received the best sum of costs/utilities included in the messages which were received most recently from its neighboring function-nodes.

## 4.2 Local Search with Asymmetric Constraints

Let us start the discussion of local search algorithms for ADCOPs by demonstrating the shortcomings of existing local search methods.

Consider again the problem described in Figure 1. Assuming each agent is only aware of the left (agent $A_1$) or right ($A_2$) value in the matrix, standard DCOP local search algorithms such as DSA and MGM can be applied to this problem. In DSA, for example, agents only consider their personal gain, or improvement, and as a result change values according to their local state. A similar situation exists with respect to MGM. However, the maximum change that is reported by agents running MGM does not necessarily imply an improvement to neighbors as well.

---

2. We preserve the terminology of Farinelli et al. (2008) and call constraint representing nodes in the factor graph "function-nodes".

The asymmetric structure of constraints alters the algorithms' behavior. For example, while DSA and MGM converge to local optima on standard DCOPs, this is not true for ADCOPs. In local search, agents continuously attempt to change their value assignment if an improving value assignment exists. When no such value assignment is found by any of the agents the state of the system as a whole is said to be stable. This state is not necessarily a local optimum when asymmetric payoffs are considered. A change of an assignment by an agent may increase its own local cost, but due to asymmetry this change can also result in an overall lower cost to the system as a whole! On the other hand, such stable solutions comply with the definition of *Nash Equilibria* (NE) – no unilateral change by any single agent can improve its state. For similar reasons, MGM in ADCOPs looses its important monotonicity property (Maheswaran et al., 2004a). Agents sending their maximal possible improvement to the current state to their neighbors can actually consider a change that would cause a deterioration of the state of their neighbors and of the global state.

Nash Equilibria do not necessarily coincide with the optimum of a global objective function. In the well known example of the prisoners' dilemma, when maximizing the gain of participants, the globally worst solution is the only NE. It is important to note that NEs do not exist in every asymmetric problem and even in the presence of a NE, it is possible that neither DSA nor MGM will converge to it. Thus, the convergence prediction for DCOPs made by Chapman et al. (2008) does not apply in the case of ADCOPs.

One may attempt to run existing local search algorithms on the derived PEAV-DCOP of the asymmetric cost problem (e.g., Figure 2). However, the PEAV formulation significantly reduces the usefulness of standard local search algorithms.

PEAV includes hard equality constraints for any pair of variables which are a variable in the original DCOP and its duplication. Consider a global assignment to the problem that does not violate any of these constraints. Any attempt of an agent to replace a value assignment to one of its variables will result in a violation of a hard constraint. Thus, the representation of any assignment for the original DCOP in the PEAV representation forms a local optimum.

Another way to describe this phenomenon is by pointing out that the PEAV formulation generates new local optima, and thus, implicitly, new NEs. The new local optima can be easily observed on the main diagonal of the cost matrix of Figure 2. When considering and analyzing the personal costs of each agent from this interaction one sees that these correspond to NEs. This implies that the PEAV formulation of a given problem produces new stable points (local optima)! In the case of the example in Figure 2, four new NEs are generated where originally there were none.

The PEAV formulation does not solve the problems of any of the standard local search algorithms. In the case of DSA, an agent only considers the current assignments of its neighbors. In the case of MGM, every change to a variable that would generate an inequality would not be considered as a maximal reduction.

---

**Algorithm 6** ACLS

---

1: $value \leftarrow$ ChooseRandomValue()
2: **while** no termination condition is met **do**
3:    send $value$ to neighbors
4:    collect neighbors' values
5:    $IMP\_SET \leftarrow$ LocalReductions()
6:    $PV \leftarrow$ RandomSelectProposedValue($IMP\_SET$)
7:    send $PV$ to neighbors
8:    collect neighbors' $PVs$
9:    **foreach** neighbor $A_n$ **do**
10:      send constraint cost with $A_n$'s $PV$
11:    collect all constraint costs
12:    $cost \leftarrow$ SumOfAllConstraintsCosts() $\cdot$ $C$
13:    **if** $cost < currentState$ **do**
14:      assign with probability $p$: $value \leftarrow PV$

---

## 4.3 ADCOP Local Search Algorithms

The aforementioned shortcomings of standard local search algorithms on both the ADCOP model and the PEAV formulation call for the development of new local search algorithms specifically designed for ADCOPs. Such algorithms that attempt to incorporate information from the agent's local neighborhood and utilize it to locate high quality solutions are presented next.

### 4.3.1 ASYMMETRIC COORDINATED LOCAL SEARCH (ACLS)

The ACLS algorithm presented in Algorithm 6, attempts to combine information from each agent's surrounding in order to produce a global evaluation.

It proceeds in synchronous steps and continues running (after a random initial assignment) until a termination condition is met. At each step, an agent running ACLS begins by sending its current assignment to its neighbors and collecting assignments from them (lines 3-4). It then collects all assignments that can improve its local state (line 5). Based on this improving set a proposed assignment $PV$ is randomly picked according to the distribution of gains from each proposal (line 6). This proposal is sent to all neighbors and the neighbors' proposals are collected (lines 7,8). An agent receiving a proposal responds with the value of its side of the constraint resulting from its current assignment and the proposed assignment (lines 9-10). When all such *impact* messages arrive, the agent assesses the potential gain or loss from the assignment (lines 11-12). ACLS agents use a special coordination value, $C$, representing the amount of cooperation with their neighborhood. That is, when this constant is zero, all *impact* messages are ignored and ACLS produces results similar to those of DSA (albeit with a high overhead of network load and privacy degradation). An agent running ACLS concludes each round by committing to a change with probability $p$ (lines 13-14). The use of the probability parameter $p$ here is similar to the use of $p$ in DSA (Zhang et al., 2005). It prevents many concurrent changes in the same neighborhood that may cause thrashing.

---

**Algorithm 7** MCS-MGM

---

1: $value \leftarrow$ ChooseRandomValue()
2: **while** no termination condition is met **do**
3:     send $value$ to neighbors
4:     collect neighbors' values
5:     **foreach** neighbor $A_n$ **do**
6:         $\Delta \leftarrow$ increase due to $A_n$'s new value
7:         **if** $\Delta > A_n$'s last known $LR$ **do**
8:             send constraint cost with $A_n$'s new value
9:             change constraint cost with $A_n$'s new value to 0
10:     collect neighbors' constraint updates
11:     update constraint with each of the neighbors
12:     $LR \leftarrow$ BestPossibleLocalReduction()
13:     send $LR$ to neighbors
14:     collect neighbors' $LRs$
15:     **if** $LR > 0$ **do**
16:         **if** $LR > LRs$ of neighbors (ties broken using indexes) **do**
17:             $value \leftarrow$ the value that gives $LR$

---

### 4.3.2 Minimal Constraint Sharing MGM (MCS-MGM)

Similarly to ACLS, the MCS-MGM algorithm presented in Algorithm 7 also attempts to employ knowledge of its local neighborhood to achieve a better gain to its surroundings.

The MCS-MGM algorithm also proceeds in synchronous steps and terminates according to a predefined condition. Each step consists of three different interaction phases. An agent begins by exchanging assignments with its neighbors (lines 3-4). It then evaluates the impact of its neighbor's assignment change on its own local state. If the neighbor's assignment change degrades the current state by more than that neighbor's last known best local reduction, the constraint is passed on to the neighbor. That is, the agent sends to its neighbor its side of the constraint with the neighbor's new value, and assigns a cost of zero instead (lines 5-9). The updated constraints are gathered and the local sub-problem is slightly modified (lines 10-11). Using the new information, the agent seeks the best local reduction and sends this information to its peers (lines 12-13). As in MGM, the agents declaring the highest local reductions change their values (lines 14-17).

### 4.3.3 Guaranteed Convergence Asymmetric MGM (GCA-MGM)

A small adjustment to *MCS-MGM* can guarantee its convergence to local optima (note that it converges to local optima and not to a NE). Line 7 is replaced by:

7:         **if** $\Delta > 0$ **do**

We call the resulting algorithm Guaranteed Convergence Asymmetric MGM (GCA-MGM). The guaranteed convergence comes with a price. GCA-MGM is expected to preserve less privacy than MCS-MGM since it has a weaker condition for exchanging constraints among agents (see Section 5).

**Proposition 5** *GCA-MGM is guaranteed to converge to a local optimum in a finite number of steps.*

**Proof:** Assume that GCA-MGM does not converge. Consequently, agents are repeatedly changing their value assignments. Each change that causes an increase for some agent triggers a constraint exchange and therefore the next time this assignment change is performed, it will not cause an increase (i.e., cannot occur more than once). Thus, the number of increases in cost is bounded by the number of constraints, which is finite. After all possible increments have caused an exchange of constraints, the convergence is guaranteed as for standard MGM (Maheswaran et al., 2004b; Pearce & Tambe, 2007). □

Our experiments (Section 5) demonstrate that although MCS-MGM does not guarantee convergence, both versions converge very rapidly. This rapid convergence has a strong impact on privacy loss during the search process (Section 5.2).

## 5. Experimental Evaluation

The experimental evaluation is divided into two parts. In the first part, the performance of the complete ADCOP algorithms that were presented in Section 3 is compared to the performance of standard state-of-the-art complete DCOP algorithms when solving asymmetric problems that are represented by the PEAV formulation. Two standard measures of performance for complete algorithms are used – runtime and network load.

In the second part of the experimental evaluation, the proposed ADCOP local search algorithms are compared with state-of-the-art local search algorithms for solving DCOPs. The focus of the evaluation of incomplete algorithms is on the quality of the solution they produce in a given limited run time. As will be demonstrated by the experimental evaluation, the proposed ADCOP local search algorithms converge to a high quality solution, in contrast to standard incomplete methods which do not converge.

The privacy loss incurred by running each of the ADCOP algorithms (both complete and local search) is also of great interest. Following Greenstadt et al. (2006) and Brito et al. (2009), the privacy loss is measured in terms of entropy.

Several domains of evaluation are used. The first domain, random asymmetric Max-DisCSP, is used to evaluate both complete and local search algorithms. Max-DisCSP is a subclass of DCOP in which all constraint costs are equal to one (Modi et al., 2005). Max-CSPs are commonly used in experimental evaluations of constraint optimization problems (COPs) (Larrosa & Schiex, 2004) and of Distributed COPs (Gershman et al., 2009). Max-DisCSPs are classified by the number of agents $n$ (assuming each holds exactly one variable), the size of the domains $k$, the probability of a constraint among any pair of variables $p_1$, and the probability for the occurrence of a violation (a non-zero cost) among two value assignments $p_2$. In our formulation we consider the constraint tightness $p_2$ as the average fraction of forbidden value pairs, as viewed by each agent involved in a given constraint. This implies that some pairs allowed by one of the involved agents are disallowed by the other, and vice versa. As a result, the fraction of cost-inflicting pairs is greater than the $p_2$ value. We refer to this fraction as $p_{2-eff}$. Its expected value is:

$$p_{2-eff} = 1 - (1 - p_2)^2 \tag{3}$$

The second evaluation domain is that of random graphical games (Kearns et al., 2001; Nisan, Roughgarden, Tardos, & Vazirani, 2007; Maheswaran et al., 2004a). In these problems, each constraint between two agents represents a local randomly generated game. In these local interactions, each constrained agent is assigned a cost for each joint action (value assignment pair) of the two constrained agents. The goal of the agents is to reach a globally minimal cost assignment. In the evaluation of ADCOP local search algorithms, general-form graphical games, as well as scale-free networks, were used. More details on these domains are given in Section 5.2.

In order to evaluate the runtime performance of the algorithms, we measure Non-Concurrent Logical Operations (NCLOs) (cf., Netzer, Grubshtein, & Meisels, 2012). This measure, which is based on the notion of atomic operations (Gershman, Zivan, Grinshpoun, Grubshtein, & Meisels, 2008), is a generalization of NCCCs (Zivan & Meisels, 2006). NCLOs enable the comparison of runtime performance with algorithms that their basic operation is not necessarily a Constraint Check, such as ODPOP (Petcu & Faltings, 2006).

## 5.1 Evaluation of ADCOP Complete Search Algorithms

The evaluation of the ADCOP complete search algorithms consists of three experiment settings – two settings of random asymmetric Max-DisCSPs, followed by a graphical games setting. 50 random instances were generated for each set of experiments and the results were averaged over these 50 instances.

In the first set of experiments, random asymmetric Max-DisCSPs with 10 agents ($n = 10$), 10 values ($k = 10$), constraint density $p_1 = 0.4$, and varying constraint tightness $0.1 \leq p_2 \leq 0.9$ were generated. Figures 7 and 8 present the runtime and network load results of running the proposed ADCOP algorithms. Figure 7 presents the results for the mean number of NCLOs. In these algorithms, we measure Constraint Checks, i.e., NCCCs. The results show that SyncABB outperforms ATWB for most of the problems. This suggests that asynchronicity actually impairs performance in ADCOPs, in contrast to the symmetric case of DCOPs. Verifying the bound by sending the CPA backwards sequentially before resuming the search is more efficient than continuing with the search while bounds are checked asynchronously. As problems become tighter, the effectiveness of two-way bounding increases. For very tight problems ($p_2 = 0.9$) the performance of ATWB is very close to that of SyncABB.

The results for the total number of sent messages are presented in Figure 8. As expected the network load of ATWB is higher than that of SyncABB due to the overhead of the added **BOUND_CPA** and **ESTIMATE** messages.

In the second set of experiments, asymmetric Max-DisCSPs with 6 agents ($n = 6$), 6 values ($k = 6$), constraint density $p_1 = 0.5$, and varying constraint tightness $0.1 \leq p_2 \leq 0.9$ were randomly generated. This value of $p_1$ (0.5) was chosen to ensure that most of the generated constraint graphs are connected, which is important for the faithful evaluation of algorithms that employ a pseudo-tree (BnB-ADOPT and ODPOP). Applying the PEAV formulation, an equivalent set of symmetric problems was also generated, enabling the comparison of performance to existing DCOP algorithms. The symmetric formulation was much larger in terms of variables and the number of constraints than the corresponding ADCOPs forcing the use of relatively small problems.

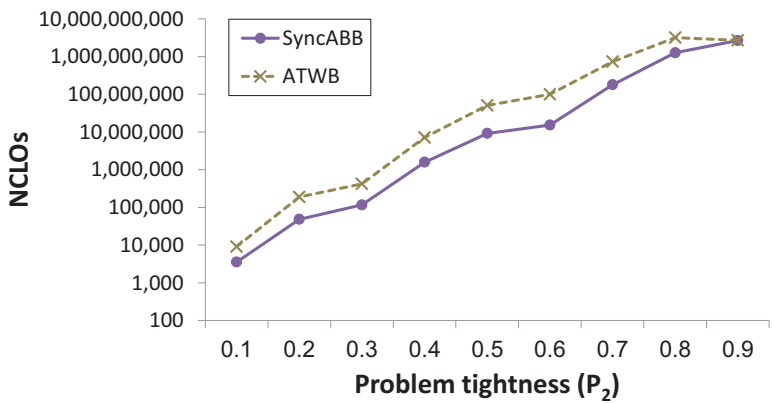

Figure 7: Mean NCLOs of complete algorithms – asymmetric Max-DisCSPs (10 agents).

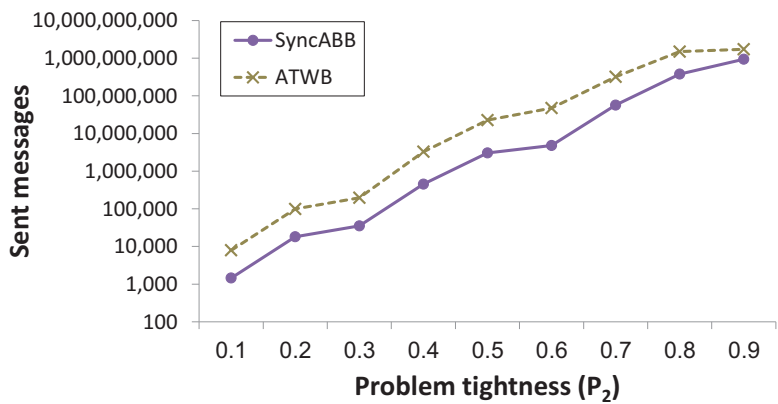

Figure 8: Sent messages in complete algorithms – asymmetric Max-DisCSPs (10 agents).

In this setup, the runtime and network load of six algorithms were compared. These included the proposed ADCOP algorithms SyncABB and ATWB, as well as several state-of-the-art DCOP algorithms (solving the symmetric problems) – SyncBB (Hirayama & Yokoo, 1997), AFB (Gershman et al., 2009), BnB-ADOPT (Yeoh et al., 2010), and ODPOP (Petcu & Faltings, 2006).

Figures 9 and 10 present the results of the second setup. Unlike the other algorithms, the main computational operation in ODPOP is the comparison of combinations of assignments sent to each computing agent by its offspring in the pseudo tree (Petcu & Faltings, 2006). Figure 9 presents the runtime in terms of NCLOs of all six algorithms. The AD-COP algorithms show lower runtime by several orders of magnitude compared to SyncBB, AFB, and ODPOP. ODPOP ran out of heap memory (heap set to 2GB) in relatively tight problems ($p_2 \geq 0.5$). This is not surprising, since the memory that ODPOP requires is exponential in the induced width of the constraint graph (Petcu & Faltings, 2006).

The PEAV formulation results in very sparse problems. The PEAV representation of an asymmetric problem with 10 variables and $p_1 = 0.4$ includes 46 variables and its density is $p_1 = 0.07$. Consequently, BnB-ADOPT, which is very efficient in solving sparse problems, does well and displays runtime performance comparable to the ADCOP algorithms.

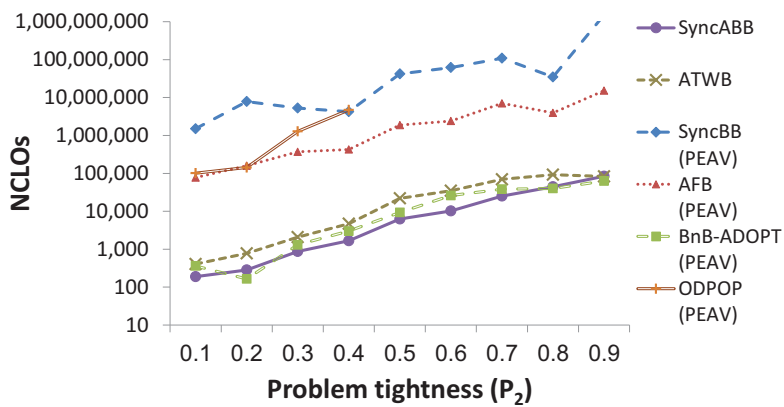

Figure 9: Mean NCLOs of complete algorithms – asymmetric Max-DisCSPs (6 agents).

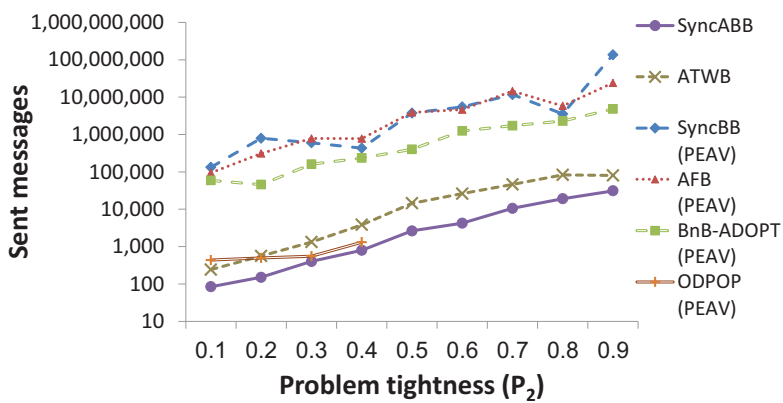

Figure 10: Sent messages in complete algorithms – asymmetric Max-DisCSPs (6 agents).

However, as demonstrated in Figure 10, it incurs a high network load. In fact, none of the algorithms solving symmetric DCOPs (PEAV) managed to complete the first set of experiments (with 10 agents) due to their high runtime and/or network load.

In the third set of experiments, graphical games with 6 agents ($n = 6$), 6 values ($k = 6$), and varying node degree were randomly generated. The average node degrees used spanned from 2.5 to 5 (which indicates a complete graph). Each constraint matrix included 50% zeroes, while the rest of the assignment pair values had random values in the range [0..9]. This setting is particulary interesting as it shows how varying density (node degree) effects the performamce of the algorithms.

Figures 11 and 12 present the results of the third setup in linear scale. The results indicate that increased density (node degree) has minor effect on the performance of ADCOP algorithms. On the other hand, as the density of the problems increases, the performance of BnB-ADOPT is drastically impaired. As suspected, the number of sent messages is especially high in BnB-ADOPT. The other DCOP algorithms (using the PEAV formulation) were not able to complete the graphical games set of experiments.

The average privacy loss incurred by running each of the complete ADCOP algorithms in the first problem setup is presented in Figure 13. Following Brito et al. (2009), the privacy results were calculated by comparing the percentage of lost entropy at the end of the computation to the initial entropy. The results clearly indicate that ATWB has a higher degree of average privacy preservation when compared to SyncABB. This is not surprising when one considers that in ATWB information is revealed only in a single direction, while SyncABB decreases entropy by agents in both lower and higher priority.

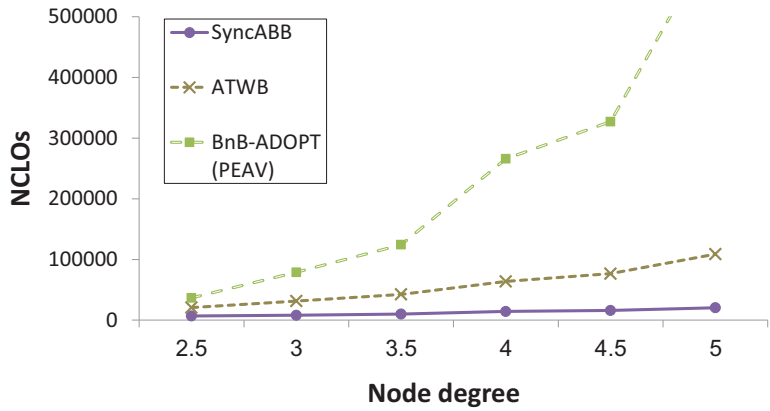

Figure 11: Mean NCLOs of complete algorithms – graphical games.

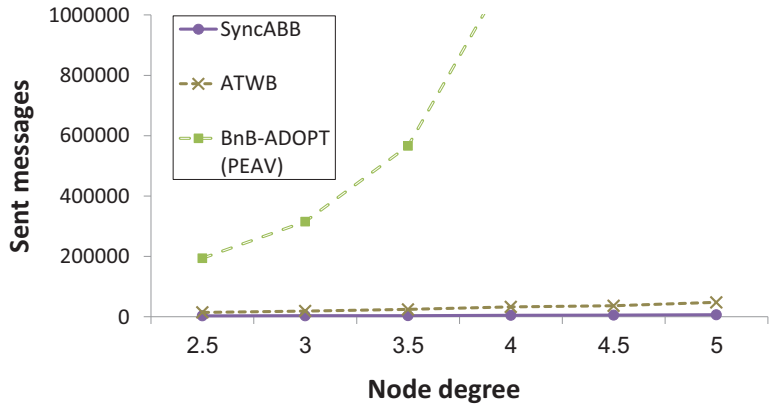

Figure 12: Sent messages in complete algorithms – graphical games.

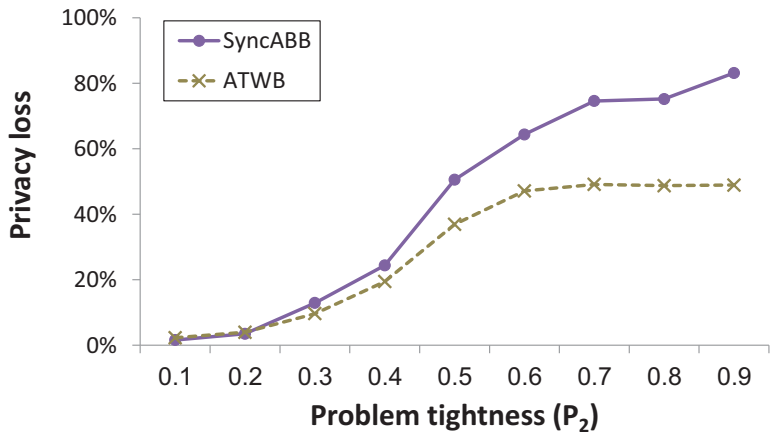

Figure 13: Privacy loss of complete algorithms – asymmetric Max-DisCSPs (10 agents).

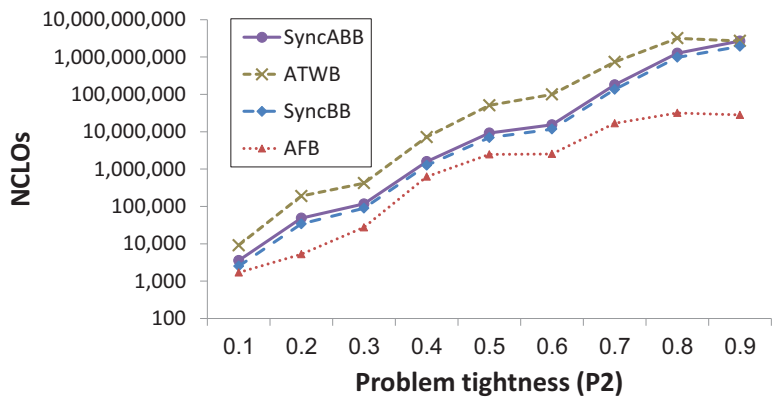

Figure 14: Mean NCLOs of complete algorithms – asymmetric Max-DisCSPs (10 agents).

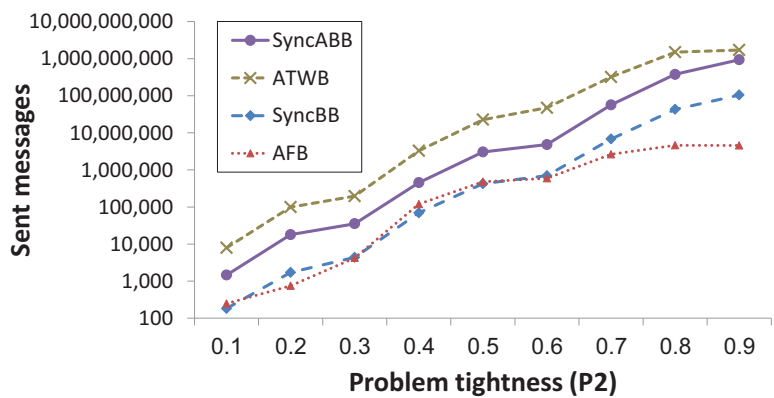

Figure 15: Sent messages in complete algorithms – asymmetric Max-DisCSPs (10 agents).

An additional method for solving problems with asymmetric constraints is to aggregate both sides of all constraints and simply run a symmetric DCOP algorithm (Section 2.3.1). The clear downside of this alternative is that all private constraint information is disclosed a-priory. It is interesting to investigate how the attempt to keep constraint information private (by using ADCOPs) affects performance. The following figures present experiments in which the performance of ADCOP algorithms, SyncABB and ATWB, is compared to the performance of their symmetric variants when using constraints disclosure. In Figures 14 and 15 the four algorithms were compared when running on the first problem setup. In the comparison between SyncABB and symmetric SyncBB one can observe that SyncABB performs only about 30%-40% more NCLOs than SyncBB, while the network load of SyncABB is just bellow one order of magnitude higher than that of SyncBB. These results indicate that the impact on performance when using ADCOPs for preserving portions of private information is reasonable. Different outcomes follow the comparison between ATWB and symmetric AFB, as the number of NCLOs in ATWB is almost two orders of magnitude

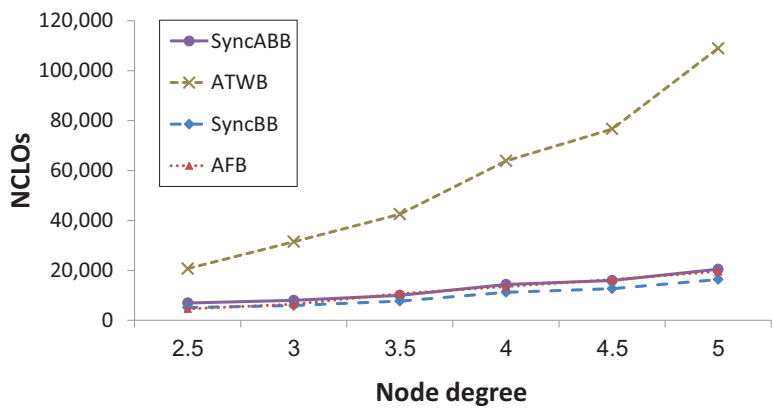

Figure 16: Mean NCLOs of complete algorithms – graphical games.

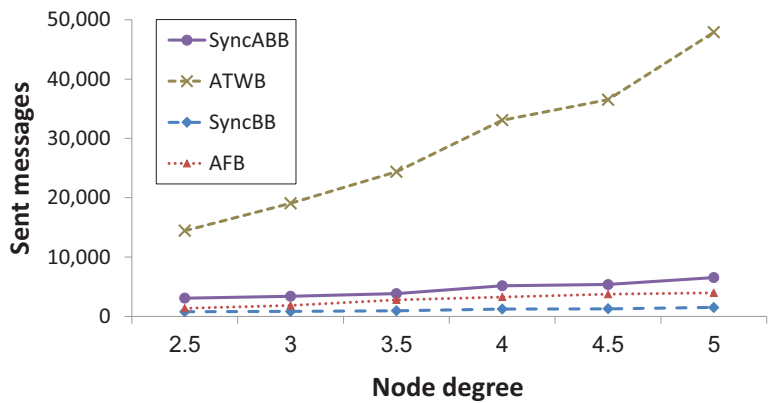

Figure 17: Sent messages in complete algorithms – graphical games.

larger than that of AFB in tight problems. The gap is even greater when considering network load.

In the graphical games setup, where simple branch and bound is more effective (due to the variance in costs), the performance impairment of the proposed ADCOP algorithms is even further refined. Figure 16 shows that symmetric SyncBB outperforms SyncABB in terms of runtime by just 25% in the higher density problems. The gap between ATWB and symmetric AFB is also relatively low, as ATWB is about 5 times faster than AFB. Moreover, the variance in costs renders that SyncABB performs less NCLOs than symmetric AFB in the higher density problems. The gaps in network load are slightly larger, as is shown in Figure 17.

While ADCOP algorithms prevent a-priori loss of all private information, the results given in Figure 13 reveal that for relatively hard problems, a major part of the private information is revealed after all. In an additional experiment, the extent of privacy loss was limited by stopping the search process whenever the amount of private information that some agent gained has passed a predefined threshold. Figures 18 and 19 present the outcomes of this limitation in terms of solution quality (distance in cost from the optimal solution) for problems from the first setup with $p_2 = 0.3$ and $p_2 = 0.5$, respectively. These specific $p_2$ values were chosen, since $p_2 = 0.3$ represents rather tight but still satisfiable problems, while $p_2 = 0.5$ represents harder problems that are mostly unsatisfiable. For $p_2 = 0.3$ the solution cost drastically reduces as the threshold is raised and with a maximal private information gain of 60% the algorithms reach the optimal solution. The average privacy loss in this case is considerably lower and is around 10% in both algorithms (see Figure 13). For $p_2 = 0.5$ some agent usually gains all the private information in order to reach the optimal solution, but as can be seen in Figure 13, the average privacy loss of all agents is much lower.

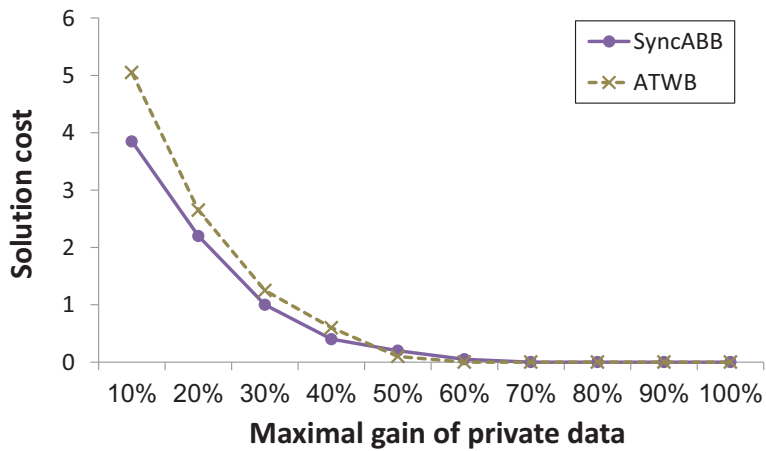

Figure 18: Solution quality when limiting the maximal gain of private information in complete algorithms – asymmetric Max-DisCSPs (10 agents, $p_1 = 0.4$, $p_2 = 0.3$).

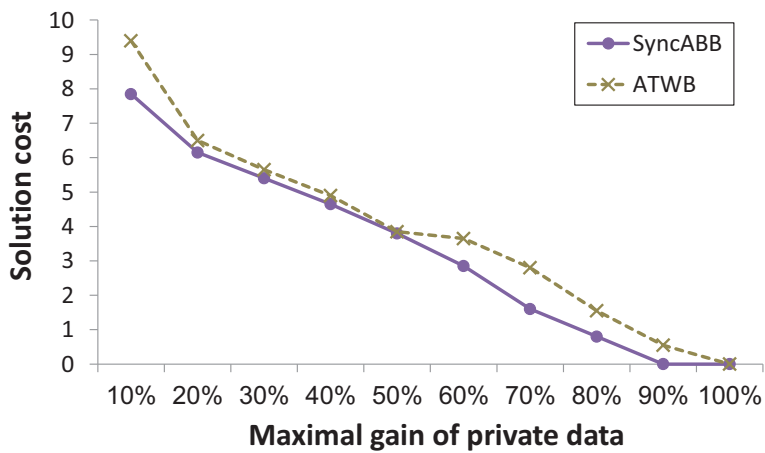

Figure 19: Solution quality when limiting the maximal gain of private information in complete algorithms – asymmetric Max-DisCSPs (10 agents, $p_1 = 0.4$, $p_2 = 0.5$).

It is interesting that for most threshold values the maximal gain of private information in ATWB is higher than in SyncABB, while it was shown that the average privacy loss of ATWB is lower (Figure 13). The reason for this phenomenon is that in ATWB agents only gain private information of their predecessors, so commonly the last agents in the order gain more information, while the first agent gains none. The result is that on average an agent running ATWB looses less privacy than an agent running SyncABB, but in SyncABB the privacy loss is better distributed among the agents. This last experiment illustrates the clear tradeoff between privacy loss and solution quality when running ADCOP algorithms with a privacy gain threshold.

## 5.2 Evaluation of ADCOP Local Search Algorithms

The introduced local search algorithms are evaluated over three domains – asymmetric Max-DisCSPs, general-form graphical games, and scale-free networks. For each type of problem, 500 different problem instances with 200 agents and a domain size of 10 values per agent were generated. The presented results are the average over the 500 solutions obtained by the algorithms for these instances. Asymmetric Max-DisCSPs were generated with an average of 10 neighbors per agent (density parameter $p_1 = 0.05$) and tightness parameter $p_2 = 0.7$. The rest of the details are as described above for the asymmetric Max-DisCSPs generated to evaluate complete algorithms. In the general-form graphical games agents had an average of 5 neighbors each, i.e., agents were connected by an asymmetric constraint with a probability value $p = 0.025$ (Erdös-Rényi graphs) (Erdös & Rényi, 1960). In the scale-free networks domain, graphs were constructed by following the Barabási-Albert model (Jackson, 2008). In the two latter setups the constraint costs were selected in the range [0..100]. Costs were separately selected in a non uniform manner – the cost for an agent was 0 with probability 0.35 and uniformly selected in the range [1..100] with probability 0.65. This structure ensures that an improving assignment change to one agent can increase the cost incurred on its neighbors (cf. Equation 3).

The scale-free networks were built using the Barabási-Albert model. An initial set of 10 agents was randomly selected and connected. At each iteration of the Barabási-Albert procedure an agent was added and connected to 4 other agents with a probability that is proportional to the number of links that the existing agents already have.

A large number of algorithms were examined. These include DSA, MGM, MGM-2, Max-Sum, ACLS, MCS-MGM, and GCA-MGM. These algorithms were executed for a maximum of 200 cycles, where a cycle includes all actions between two consecutive *value* messages sent by the same agent (Max-Sum cycles include both messages from function-nodes to variable-nodes and vise versa).

Figure 20 presents the average solution quality for asymmetric Max-DisCSPs for each of the algorithms as a function of cycles. MCS-MGM produced the highest quality results and after 200 cycles the best algorithms were MCS-MGM, MGM-2, ACLS, and GCA-MGM, where the performance of the two latter algorithms was similar. All three ADCOP algorithms demonstrated very fast convergence. The highest costs (e.g., worst solutions) were reported by Max-Sum, which did not explore much of the search space and its reported incurred cost was significantly higher than that reported by other algorithms. Consequently, the results of Max-Sum were left out of the plots to enable a better view of the overall performance of the remaining algorithms (and a correct scale). Surprisingly, DSA produced solutions of significantly lower quality than MGM. This is in contrast to its performance

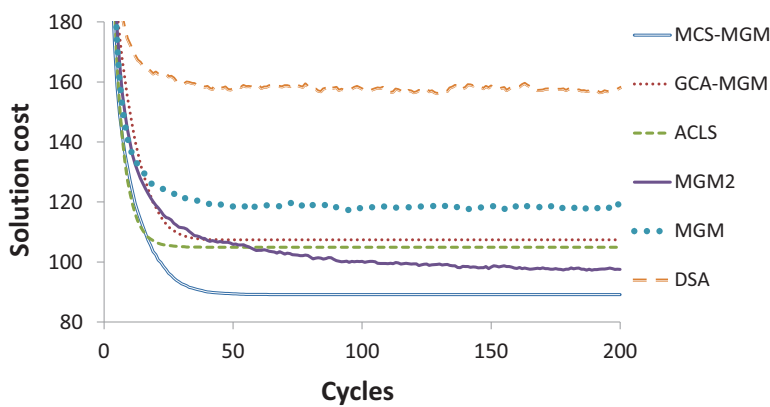

Figure 20: Solution quality of local search algorithms – asymmetric Max-DisCSPs

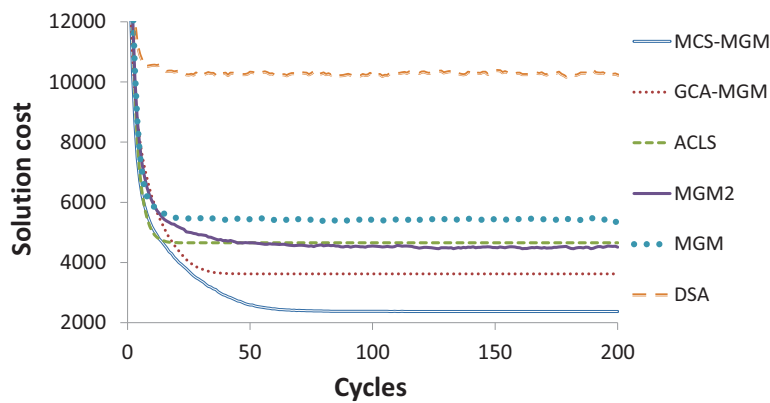

Figure 21: Solution quality of local search algorithms – Erdös-Rényi graphs

when solving symmetric problems, where DSA is known to produce higher quality solutions than MGM.

Figures 21 and 22 present similar results for the average solution quality in graphical games and scale-free networks, respectively. It is notable that MCS-MGM dominates in all problem scenarios. GCA-MGM produces better results than MGM-2 in graphical games, while their results are of similar quality in scale-free networks. DSA and Max-Sum produced low quality results in these problem scenarios as well.

It is worth noting that in MGM-2, an agent optimizing for itself and another agent can cause an increase in the valuation of the proposed alternative state for neighboring agents of both. As a result, agents optimizing for different pairs can generate loops of assignment changes just as described for MGM. Thus, the increase in the size of the group of agents considered by the optimizing agent is not sufficient to ensure convergence. A similar phenomenon, where MGM-2 can eventually fail to provide higher quality solutions than even MGM, was reported in the presence of uncertainty (Taylor, Jain, Tandon, & Tambe, 2009).

In the problem scenarios evaluated, ACLS produced results of lower quality than MCS-MGM and GCA-MGM (except for the similar results in asymmetric Max-DisCSPs). How-ever, it is also observable that ACLS is the fastest algorithm to converge in all three problem scenarios.

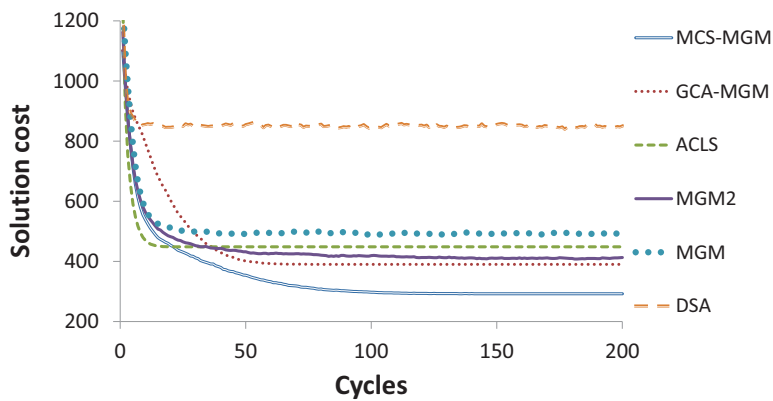

Figure 22: Solution quality of local search algorithms – scale-free networks

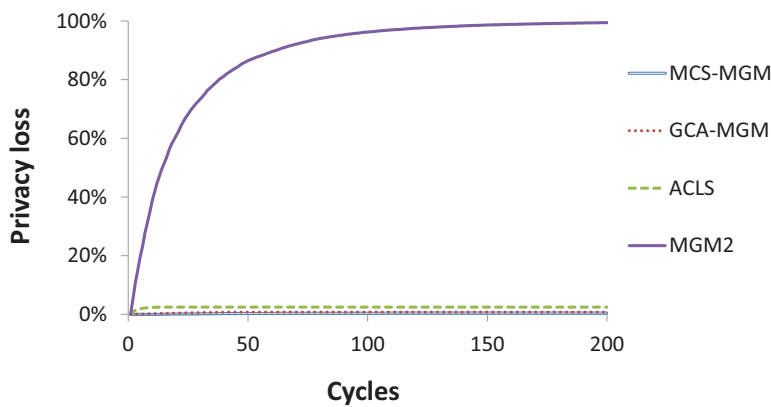

Figure 23: Privacy loss of local search algorithms – asymmetric Max-DisCSPs

The above results indicate that the cooperation inherent to the three proposed algorithms, ACLS, MCS-MGM, and GCA-MGM, renders these algorithms better suite the asymmetric case than most other algorithms. However, despite the lower costs attributed with their cooperation, the nature of the agents requires some revelation of private information. Thus, it is important to asses the privacy loss resulting from the coordination of agents, in contrast to standard local search (1-opt) algorithms which preserve a high level of privacy (Greenstadt, 2008). To measure the overall loss of privacy in our system of agents, one needs to aggregate the number of revealed constraint parts by each agent (Greenstadt et al., 2006; Greenstadt, 2008).

In ACLS, a fraction of the constraint is revealed in line 10 (Algorithm 6), while MCS-MGM and GCA-MGM reveal constraint information in lines 8 and 9 (Algorithm 7). Another algorithm that attempts to coordinate joint moves is MGM-2 (Maheswaran et al., 2004a), in which *offerer* agents propose several improving assignments along with their costs to one of their peers, which respond with the lowest improving cost incurred on them. Thus, MGM-2 agents reveal a much larger fraction of the constraint in every interaction.

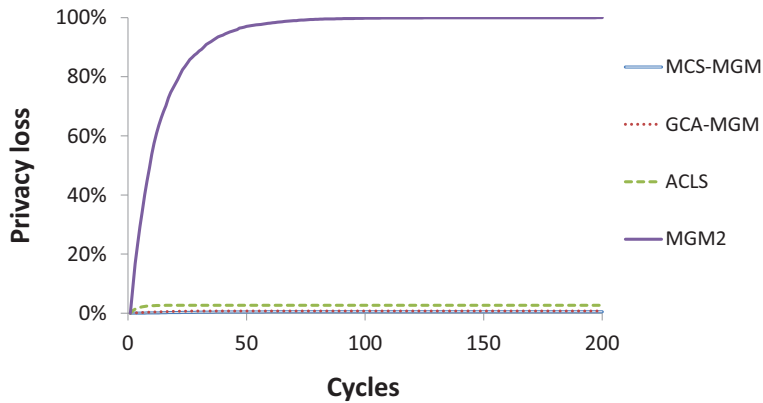

Figure 24: Privacy loss of local search algorithms – Erdös-Rényi graphs

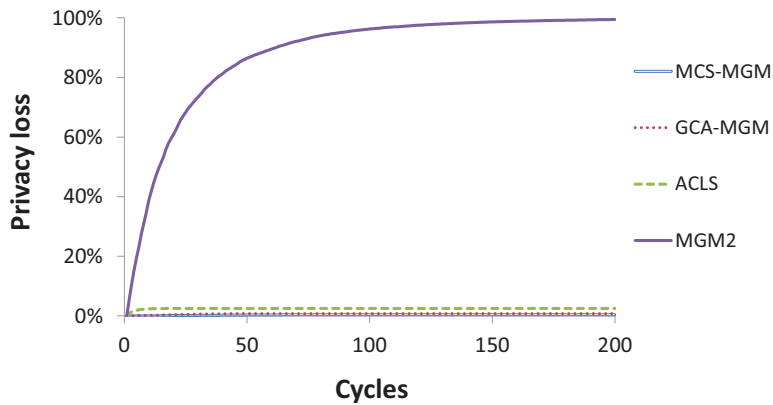

Figure 25: Privacy loss of local search algorithms – scale-free networks

Figures 23, 24, and 25 present the privacy loss measurements. Agents running MGM-2 reveal most of their problem structure, while the other algorithms maintain a substantially higher degree of constraint privacy. In all three problem scenarios it is apparent that the privacy loss of the proposed ADCOP algorithms is negligible compared to the privacy loss of MGM-2. Furthermore, the privacy of MGM-2 decreases throughout the 200 cycles of the algorithm (although the privacy loss function has a notable concave structure), while ACLS, MCS-MGM, and GCA-MGM lose a very small amount of privacy in the first few iterations, and do not endure additional privacy loss. These results indicate that quick convergence to a solution may have a substantial impact on privacy loss.

## 6. Conclusions

Many problems that are distributed by nature include agents which have different valuations of the possible states of the world. For distributed constraint optimization problems the constrained agents may have different costs assigned to valued constraints. The present paper proposes the Asymmetric Distributed Constraint Optimization Problems (ADCOP) model, which captures the inherent asymmetry of distributed problems in a natural and ef-ficient way. The proposed ADCOP model represents private gains without revealing private information a-priori. Instead, agents reveal only the information which is necessary during the distributed search for a solution. This is in contrast to alternative DCOP formulations that either centralize constraints (resulting in privacy loss and possibly heavy network load), or change the problem into a more complex structure (PEAV).

The algorithmic impact of introducing the new framework was discussed, as well as the applicability of existing DCOP algorithms. Several novel algorithms were proposed – some for complete search and others for local search.

When considering complete search, the proposed complete ADCOP algorithms eliminate the need to extend the problem by using the PEAV model. Furthermore, they behave well for the whole range of problem difficulty. Two of the complete ADCOP algorithms showed superior performance (runtime as well as network load) when compared to the leading (symmetric) DCOP algorithms that use the PEAV representation. Another alternative is to aggregate both sides of all constraints and simply run a symmetric DCOP algorithm. However, this method leads to an a-priori disclosure of all private constraint information, while only achieving moderately better run-time performance than the respective ADCOP algorithms. The results indicate that the synchronous algorithm (SyncABB) outperforms the asynchronous algorithm (ATWB) in most cases. SyncABB usually performs less NC-LOs, sends significantly less messages, and leads to a better distribution of privacy loss between the agents. In contrast, the average privacy loss in ATWB is lower.

In the proposed ADCOP local search algorithms the agents cooperate and perform search in their local neighborhood, instead of maximizing their own gain. A proof that one of the algorithms, GCA-MGM, is guaranteed to converge to a local optimum, was presented. The PEAV representation cannot be used in combination with existing local search algorithms, since any assignment which does not violate hard constraints is a local optimum that PEAV generates. However, existing local search algorithms can be used in combination with the proposed ADCOP model. Nevertheless, an empirical evaluation demonstrated that the new ADCOP algorithms consistently find higher quality solutions, and do so with a high degree of privacy preservation. It turns out that their fast convergence strongly limits the amount of privacy lost.